\pdfoutput=1
\documentclass[11pt]{article}

\usepackage{ACL2023}
\usepackage{times}
\usepackage{latexsym}
\usepackage{graphicx}
\usepackage{makecell}
\usepackage[section]{placeins}
\usepackage{subfigure}
\usepackage{makecell}
\usepackage{xcolor}
\usepackage{amsmath, bm}
\usepackage[T1]{fontenc}
\usepackage[utf8]{inputenc}

\usepackage{microtype}

\usepackage{inconsolata}

\title{An Effective Deployment of Contrastive Learning in Multi-label Text Classification}

\author{Nankai Lin\textsuperscript{1}, Guanqiu Qin\textsuperscript{1}, Jigang Wang\textsuperscript{1}, Dong Zhou\textsuperscript{2 *} \and Aimin Yang\textsuperscript{1,3 *} \\
    \textsuperscript{1} School of Computer Science and Technology, Guangdong University of Technology, \\ Guangzhou, Guangdong, 510006, China \\
    \textsuperscript{2} School of Information Science and Technology, Guangdong University of Foreign Studies, \\ Guangzhou, Guangdong, 510006, China \\
    \textsuperscript{3} School of Computer Science and Intelligence Education, Lingnan Normal University, \\ Zhanjiang 524000, Guangdong, China}
\begin{document}
\maketitle
\begin{abstract}
The effectiveness of contrastive learning technology in natural language processing tasks is yet to be explored and analyzed. How to construct positive and negative samples correctly and reasonably is the core challenge of contrastive learning. It is even harder to discover contrastive objects in multi-label text classification tasks. There are very few contrastive losses proposed previously. In this paper, we investigate the problem from a different angle by proposing five novel contrastive losses for multi-label text classification tasks. These are Strict Contrastive Loss (SCL), Intra-label Contrastive Loss (ICL), Jaccard Similarity Contrastive Loss (JSCL), Jaccard Similarity Probability Contrastive Loss (JSPCL), and Stepwise Label Contrastive Loss (SLCL). We explore the effectiveness of contrastive learning for multi-label text classification tasks by the employment of these novel losses and provide a set of baseline models for deploying contrastive learning techniques on specific tasks. We further perform an interpretable analysis of our approach to show how different components of contrastive learning losses play their roles. The experimental results show that our proposed contrastive losses can bring improvement to multi-label text classification tasks. Our work also explores how contrastive learning should be adapted for multi-label text classification tasks. 
\end{abstract}

\renewcommand{\thefootnote}{\fnsymbol{footnote}}
\footnotetext[1]{Corresponding Author. E-mail: dongzhou@gdufs.edu.cn, amyang@gdut.edu.cn.}
\renewcommand{\thefootnote}{\arabic{footnote}}

\renewcommand{\floatpagefraction}{0.9}

\section{Introduction}

Multi-label text classification is an important branch of text classification technology \cite{chalkidissogaard2022improved,zhang2022improvingmulti}. Different from binary classification tasks or multi-class classification tasks, multi-label classification tasks need to assign at least one label to a piece of text. Since the number of labels the text belongs to is not fixed, it greatly increases the difficulty of the model prediction. Specifically, the uncertainty in the number of labels poses two challenges to the training of multi-label text classification models: the output logic of the model and the semantic representation space of the model. In recent years, most multi-label text classification research has focused on designing better output logic to solve the uncertainty of the number of labels, such as transforming the multi-label text classification problem into a multi-task problem \cite{LIN2022103097}. However, for another challenge, how to construct a better semantic representation space for multi-label text classification models, little research attention has been paid.

\begin{figure}
  \centering
  \includegraphics[width=0.45\textwidth]{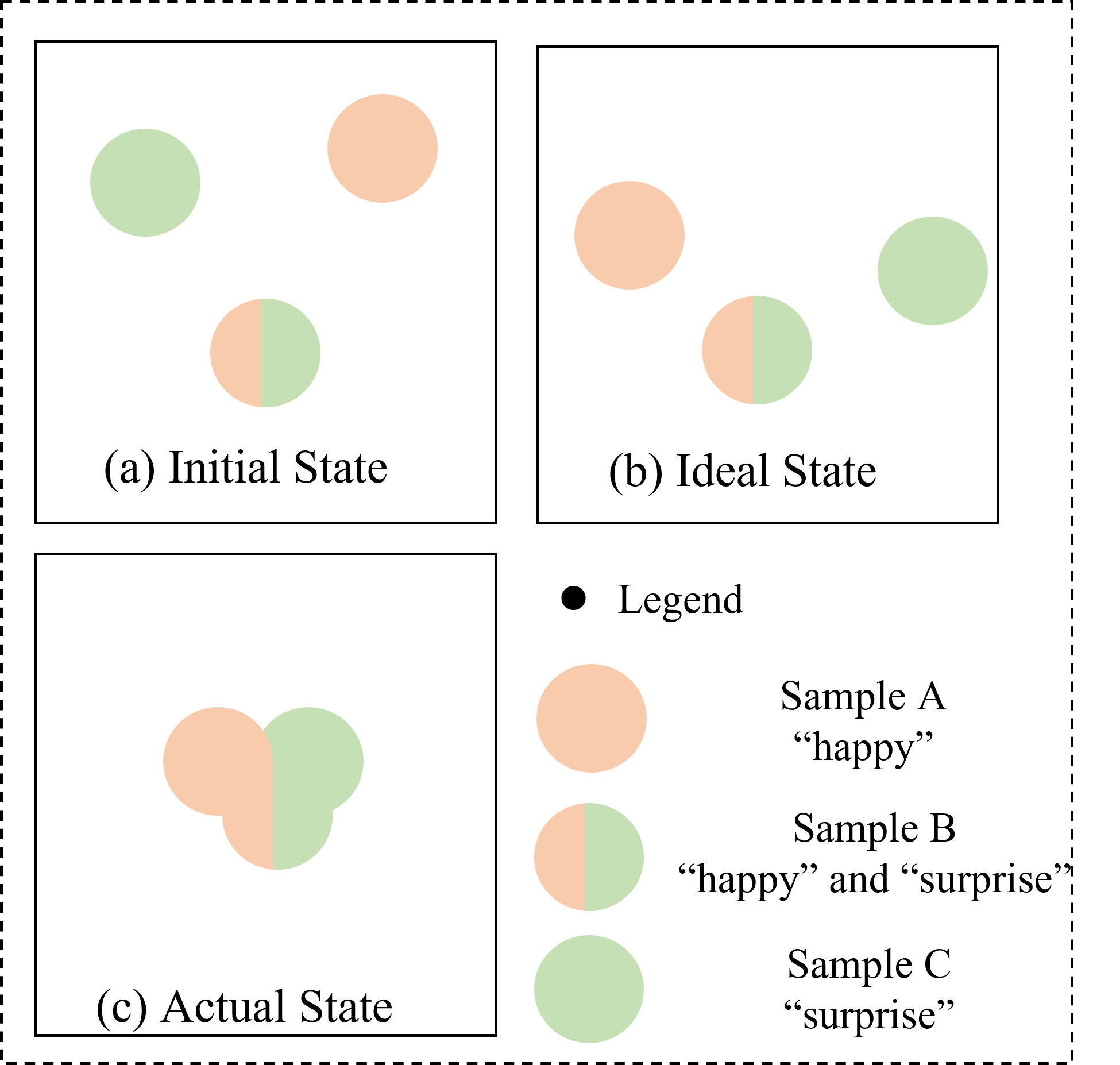}
  \caption{Example of Multi-label classification task.} 
  \label{fig:0} 
\end{figure}

The existence of multi-label samples can easily confound the semantic representation space, thereby posing a challenge in data analysis and modeling. When confronted with multi-label samples, the semantic representation space becomes susceptible to distractions, where the boundaries between different classes become blurred. This blurring effect stems from the inherent ambiguity that arises when multiple labels coexist within a single sample, causing uncertainty in the multi-label classification tasks. Take the multi-label emotion classification task as an example (shown in Figure \ref{fig:0}), in which the “happy” sample (assumed to be sample $A$) shares a label with the “happy, surprise” sample (assumed to be sample $B$), and at the same time, the “surprise” sample (assumed to be sample $C$) also shares a label with the sample $B$ (shown in Figure \ref{fig:0} (a)). Therefore, in the ideal state, the multi-label classification model assumes that sample $A$ and sample $B$ are located in a similar semantic space, and that sample $B$ and sample $C$ are located in another similar semantic space (shown in Figure \ref{fig:0} (b)). Figure \ref{fig:0} (c) shows how the samples of the “happy” category and the samples of the “surprise” category are confounded in the semantic space. This will cause sample $A$ and sample $C$ to be brought closer indirectly, even if their labels are completely different. As far as we know, the semantic representation of multi-label samples is still an open issue in the multi-label text classification task. Therefore, this paper focuses on using contrastive learning to improve the semantic representation of multi-label text classification models.

As an emerging technology, contrastive learning has achieved good performance in various fields of natural language processing \cite{NEURIPS2020_d89a66c7,gao2021simcse}. How to construct positive and negative samples correctly and reasonably is the core challenge of contrastive learning. In multi-label text classification tasks, it is a great challenge to incorporate the contrastive learning module. It is more difficult for contrastive learning to perform well in multi-label text classification tasks than in other text classification tasks because implicit information representation of multi-label text is richer in the semantic space, which makes it more difficult to define positive and negative samples. Existing studies have proposed unsupervised contrastive learning methods to improve the performance of the model on multi-label text classification tasks \cite{NEURIPS2020_d89a66c7}, and there are also working to improve supervised contrastive learning \cite{gao2021simcse}. However, the exploration of contrastive learning in multi-label text classification tasks is still very limited. 

As the typical task in multi-label text classification, multi-label emotion classification task \cite{li2022ligcn,10.1145/3394171.3413577,AMEER2023118534} and multi-label news classification task \cite{10.1007/978-3-030-86331-9_44} have received extensive attention. In this paper, we propose five contrastive losses for multi-label text classification tasks and verify the performance of our method with the multi-label emotion classification task and multi-label news classification task as the representative tasks. More specifically, they are Strict Contrastive Loss (SCL), Intra-label Contrastive Loss (ICL), Jaccard Similarity Contrastive Loss (JSCL), Jaccard Similarity Probability Contrastive Loss (JSPCL), and Stepwise Label Contrastive Loss (SLCL). These five different strategies define the positive samples and negative samples of contrastive learning from different perspectives to pull the distance among different types of samples into the semantic space. To compare the effects of the five strategies, we further conduct an interpretable analysis to investigate how the different contrastive learning methods play their roles. The experimental results show that our proposed contrastive losses can bring improvement for multi-label text classification tasks. In addition, our methods could be considered as a set of baseline models of viable contrastive learning techniques for multi-label text classification tasks. This series of contrastive learning methods are plug-and-play losses, which can be applied to any multi-label text classification model, and to a certain extent, bring effective improvements to the multi-label text classification model.

The major contributions of this paper can be summarized as follows:

(1) For multi-label text classification tasks, we propose five novel contrastive losses from different perspectives, which could be regarded as a set of baseline models of contrastive learning techniques on multi-label text classification tasks.

(2) To the best of our knowledge, this is the first work that proposes a series of contrastive learning baselines for multi-label text classification tasks. At the same time, we also explore in detail the impact of different contrastive learning settings on multi-label text classification tasks.

(3) Through interpretable analysis, we further show the effectiveness of different contrastive learning strategies in transforming the semantic representation space.

\section{Related Work}
\subsection{Multi-label Text Classification}
In the field of text classification, multi-label text classification (MLTC) is always a challenging problem \cite{LIN2022103097}. A sample of multi-label text classification consists of a text and a set of labels. There is a correlation among labels. For this, some research transforms the multi-label classification problem into the seq2seq problem and learns the potential correlation among labels with the sequence generation model \cite{NIPS2017_2eb5657d,yang2018sgm,XIAO2021107094}. \citet{yang2019deep} proposed a reinforcement learning-based seq2set framework, which can capture the correlation among tags and reduce the dependence on tag order. In addition, there is some research introducing label embedding so that the model can simultaneously learn the feature information of text and the co-occurrence information of labels. \citet{ma2021label} proposed to learn statistical label co-occurrence via GCN. LELC (Joint Learning from Label Embedding and Label Correlation) simultaneously learned labels attention and label co-occurrence matrix information \cite{LIU2021385}. \citet{zhang2021enhancing} ensembled the MLTC and the label co-occurrence task to enhance label correlation feedback.

Most dataset of MLTC has the data distribution imbalance problem: imbalance within labels, among labels, and among label-sets. The studies we have discussed above, which use label embedding, have alleviated the impact of label imbalance to some extent while learning label association. Some research solves the problem of data imbalance by resampling. For example, based on the edited nearest neighbor rule, \citet{Charte2014MLeNNAF} proposed a multi-label undersampling algorithm. They defined a measure of the differential distance between label sets in order to heuristically remove unnecessary samples during resampling. Considering the problem in terms of object functions, \citet{ridnik2021asymmetric} proposed an asymmetric loss that dynamically adjusts the asymmetry levels to balance the effect of positive and negative samples in training.

\subsection{Multi-label Emotion Classification}
Sentiment analysis \cite{7817044} is of great significance to society, economy and security. In early studies sentiment analysis  \cite{mohammad2013crowdsourcing,turney2002thumbs} is implemented based on the sentiment polarity dictionary. These methods utilize unsupervised methods such as point mutual information (PMI) to construct an emotional dictionary based on the basic emotional word set, and then calculate the emotional weight value and emotional polarity of the text according to the viewpoint words with different intensity of positive, neutral and negative emotional tendencies in the dictionary. While some studies \cite{socher2013recursive,nakov2013semeval} transform sentiment analysis into binary or mutil-classification problems, which leads to many subsequent supervised learning studies based on machine learning and neural networks.

In recent years, more and more scholars \cite{shmueli2021happy,mohammad2018semeval} regarded the sentiment analysis task as a multi-label problem, and accordingly, \citet{9490212} introduced it into multi-label sentiment analysis by adapting the focal loss and proposed a dynamic weighting method to balance each label's contribution in the training set. \citet{alhuzaliananiadou2021spanemo} transformed the problem of multi-label sentiment classification into span-prediction by means of prompt learning, and proposed a label relationship perception loss. They converted labels into tokens and inputted them into BERT together with the original input text, and used the attention module of the Transformer and the knowledge learned in the pre-train stage to learn the correlation of emotional labels. In addition to encoding labels and sentences with BERT at the same time, EduEmo \cite{ZHU2022108114} also introduced the encoder of Realformer \cite{he2021realformer} to model the association between each elementary discourse unit and sentiment labels.

\subsection{Contrastive Learning}
In recent years, contrastive learning has gradually become one of the important techniques in natural language processing and computer vision. In the field of natural language processing, contrastive learning is usually used to improve the quality of embedding representation by comparing feature vectors, bringing semantically similar and same label embeddings closer, and distancing semantically dissimilar and different label embeddings.

Contrastive learning could be divided into supervised contrastive learning and unsupervised contrastive learning. \citet{NEURIPS2020_d89a66c7} proposed a supervised contrastive learning method, which took the original label of the sample as the anchor, and made the clusters of the same label closer to each other, and the clusters of different labels far away from each other in the embedding space. To improve the sentence-level representation, SimCSE used dropout technology for unsupervised contrastive learning and natural language inference dataset for supervised contrastive learning \cite{gao2021simcse}. Some research introduced supervised contrastive learning into the pre-training process of PLMs, and experiments result on their downstream tasks showed that the performance of pre-trained models was generally improved \cite{gunel2020supervised,qin2021erica}.

\subsection{Contrastive Learning for Multi-label Text Classification}
At present, the application of contrastive learning in multi-label classification mainly focuses on image-related tasks. MulCon, an end-to-end framework for multilabel image classification, used image label-level embeddings with a multi-head attention mechanism to transform the multi-label classification problem into the binary classification problem for each label-level embedding \cite{dao2021multi}. \citet{9816043} proposed a supervised multi-label contrastive learning method for abstract visual reasoning. They reconstructed the contrastive loss function according to the multi-label problem, allowing sample pairs to contrast all labels. \citet{9880213} proposed a general hierarchical multi-label representation learning framework, which introduced hierarchical loss retention and hierarchical constraints.

However, different from the representation space of images, the implicit information representation of text is richer, which makes it more difficult to define positive samples and negative samples, and it is more difficult for contrastive learning to show good performance. Research of contrastive learning in multi-label text classification is focusing on unsupervised multi-label contrastive learning \cite{10.1007/978-3-031-03948-5_24}. What's more, \citet{su2022contrastive} attempted to improve supervised contrastive learning by using the knowledge of existing multi-label instances for supervised contrastive learning. \citet{bai2022gaussian} proposed to take the sample features as anchor samples, and take the corresponding positive labels and negative labels as positive and negative samples for supervised contrastive learning. However, the exploration of contrastive learning in multi-label text analysis tasks is still very limited.

\begin{figure}
  \centering
  \includegraphics[width=0.48\textwidth]{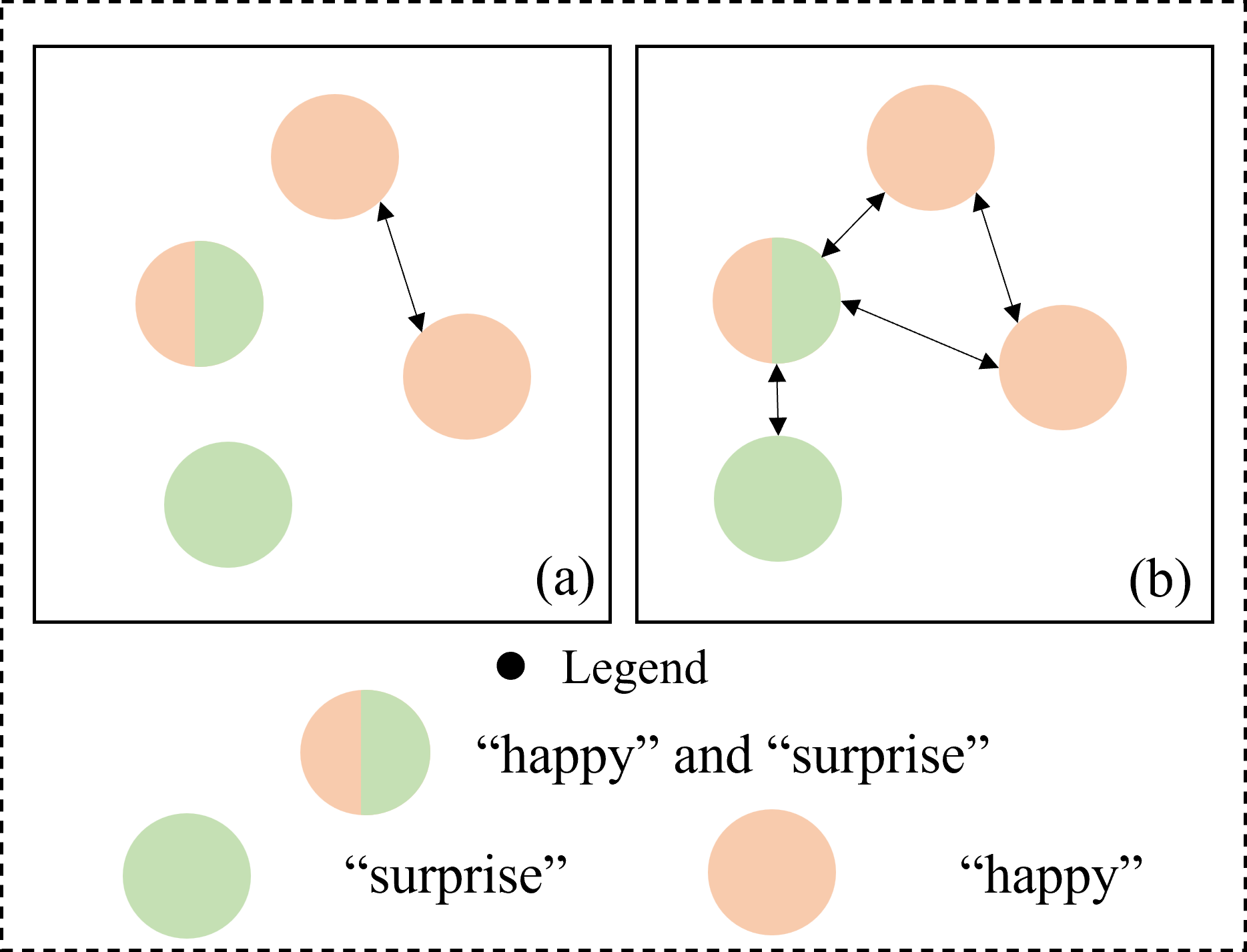}
  \caption{Example of different standards.} 
  \label{fig:1} 
\end{figure}

\section{Contrastive Loss for Multi-Label Text Classification}

In this section, we describe in detail the application of our proposed different contrastive learning methods on multi-label text classification tasks. We take the multi-label emotion classification task as an example to describe our method. It is worth noting that our proposed method can not only be applied to multi-label emotion classification tasks, but also can be applied to other multi-label text classification tasks.

Suppose a minibatch which contains $K$ samples $D=\{(X_1,Y_1 ),(X_2,Y_2 ),…,(X_K,Y_K)\}$ and $I=\{1,…,K \}$ is the sample index set. Given a sample index $i$, $X_i$ is the text sequence of samples $i$ and its label set is denoted as $Y_i$. After encoding by the multi-label text classification model $M$, we could obtain the sentence representation vector $\bm{e_i^t}$ and the emotion representation matrix $E_i^e$ of $X_i$, where $E_i^e = \{ \bm{e_{i1}},\bm{e_{i2}},…,\bm{e_{il}} \}$ and $l$ represents the total number of emotion labels. It is worth noting that the model $M$ here can be any deep learning multi-label language model. $Y_i$ is the one-hot encoding of the label, i.e. $Y_i=\{y_1,y_2,...,y_l\}$. For a given $i$-th emotion $y_i\in\{0,1\}$, $y_i$=0 means that this type of emotion does not exist in the text, and $y_i$=1 means that this type of emotion exists in the text. We further define the label prediction probability distribution of the model $M$ output as $p_i$. Contrastive learning aims to change the semantic representation space of the model. Since the multi-label classification tasks are more complex than the single-label classification tasks, the main exploration of our paper is how one can construct positive and negative samples for contrastive learning. 

When contrastive learning is applied to multi-label text classification, for an anchor, the definition of its positive samples can be diversified. For example, when a strict standard is implemented, positive samples are defined as samples with exactly the same label set (shown in figure \ref{fig:1} (a)), when a loose standard is implemented, positive samples are defined as samples with partly the same label set (shown in figure \ref{fig:1} (b)). For different positive samples, contrastive learning pulls different samples closer in semantic space for a given anchor. In the strict standard, we could find that for an anchor point, there are fewer positive samples, and samples containing some similar features cannot be pulled closer. In the loose standard, there are more positive samples for an anchor point, which may indirectly bring samples of different labels closer. Therefore, different positive and negative sample construction methods affect the optimization goal of the model. What's more, there are two different types of contrastive learning, Feature-based Contrastive Learning (FeaCL) \cite{arxiv.2204.00796} and Probability-based Contrastive Learning (ProCL) \cite{abs-2111-06021}. FeaCL uses semantic representations of sentences as the basic component to build the contrastive objective function. ProCL constructs the contrastive objective function from the perspective of probability distributions instead of semantic representations. Using different features for contrastive learning will also affect the optimization of the model. In order to explore how contrastive learning can be better applied to multi-label text classification tasks, we introduce five different contrastive learning methods SCL, ICL, JSCL, JSPCL, and SLCL, as below.

\subsection{Strictly Contrastive Loss}
As a strict standard method, SCL requires that only when the label set of the sample is exactly the same as the label set of the anchor point can it be used as a positive contrastive sample of the anchor point. Therefore, SCL does not consider samples that partially overlap with the anchor label set. In addition, SCL is also a method of FeaCL type, which uses the semantic representation of samples obtained from model encoding as the contrastive feature. In the SCL, for a given sample $i$, all other samples that share the same label set with it in the batch form the set $S=\{s:s \in I,Y_s=Y_i  \wedge s \neq i\}$. Then we could define the SCL function for each entry $i$ across the batch as

\begin{equation}
L_{SCL}= - \frac{1}{|S|} \sum_{s \in S } \log \frac
    { \exp( \frac{{sim(\bm{e_i^t},\bm{e_s^t)}}}{\tau})}
    { \sum_{k \in I \backslash \{i\}} 
    \exp( \frac{{sim(\bm{e_i^t},\bm{e_k^t)}}}{\tau})} 
\end{equation}

where $sim(\cdot)$ indicates the cosine similarity function.

\subsection{Jaccard Similarity Contrastive Loss}
SCL is a strict contrastive learning method, which only pulls the samples with the exact same label closer, while JSCL operates on the samples to different degrees according to the similarity of the labels of the samples. We use Jaccard coefficient \cite{jaccard1912distribution} to calculate the label similarity between samples. Similar to SCL, JSCL uses the semantic representation of samples obtained from model encoding as the contrastive feature. For a given sample, JSCL will zoom in as close as possible on samples with the exact same label while only slightly zooming in on samples that have some of the same labels. In the JSCL, for a given sample $i$, we could define the JSCL function across the batch as

\begin{equation}
 L_{JSCL}= - \frac{1}{|I|} \sum_{s \in I } \log \frac
    {
    \frac
    {\lvert Y_i \cap Y_s \lvert }
    {\lvert Y_i \cup Y_s \lvert } \cdot
    \exp( \frac{{sim( \bm{e_i^t}, \bm{e_s^t)}} }{\tau})}
    { \sum_{k \in I \backslash \{i\}} 
    \exp( \frac{{sim( \bm{e_i^t}, \bm{e_k^t)}} }{\tau}) }
\end{equation}

\subsection{Jaccard Similarity Probability Contrastive Loss}
\citet{abs-2111-06021} suggested that ProCL can produce more compact features than feature contrastive learning, while forcing the output probabilities to be distributed around class weights. Based on JSCL, we try to use probability for contrastive learning. In the JSPCL, for a given sample $i$, we could define the JSPCL function across the batch as

\begin{equation}
L_{JSPCL}= - \frac{1}{|I|} \sum_{s \in I } \log \frac
    {
    \frac
    {\lvert Y_i \cap Y_s \lvert }
    {\lvert Y_i \cup Y_s \lvert } \cdot
    \exp(\frac{{sim(p_i,p_s)}}{\tau})  }
    { \sum_{k \in I \backslash \{i\}} \exp(\frac{{sim(p_i,p_k)}}{\tau}) } 
\end{equation}

\subsection{Stepwise Label Contrastive Loss}
SLCL is another way to consider contrastive learning among samples with labels that are not exactly the same. The previous three contrastive learning methods mainly consider the situation when multiple emotions are considered at the same time, while SLCL considers different emotions separately, calculates the contrast loss separately, and then combines the losses of each emotion. In the JSPCL, for a given sample $i$, all other samples that share the same label $y_j$  with it in the batch form the positive sample set $S_j$. The set of positive samples under each emotion label is $S=\{ S_1,S_2,...,S_q \}$ and $q$ is the emotions’ number of sample $i$. Then we could define the SLCL function for each entry $i$ is across the batch as

\begin{small}
\begin{equation}
L_{SLCL}= - \frac{1}{q} \sum_{S_j \in S} \frac{1}{|S_j|} \sum_{s \in S_j }  \log \frac
    {
    \exp( \frac{{sim(\bm{e_i^t},\bm{e_s^t})}}{\tau})}
    { \sum_{k \in I \backslash \{i\}} \exp( \frac{{sim(\bm{e_i^t},\bm{e_k^t})}}{\tau}) } 
\end{equation}
\end{small}

\subsection{Intra-label Contrastive Loss}
Different from several other contrastive losses to narrow the semantic representation of samples with the same labels, ICL aims to make multiple emotional representations existing in the same sample closer. That is, ICL narrows the distance among emotional representations, while not narrowing the distance among sample representations. In the ICL, for a given sample $i$ and the indexes of $i$’s emotion $I_Y  = \{1,...,l\}$, we could define the ICL function for the $j$-th emotion of each entry $i$ as

\begin{small}
\begin{equation} 
L_{ICL_j}= - \frac{1}{|I_Y|} \sum_{s \in I_Y } \log \frac
    {
    \exp( \frac{{sim(\bm{e_{ij}},\bm{e_{is}})}}{\tau})}
    { \sum_{k \in I_Y \backslash \{j\}} \exp(
    \frac{{sim(\bm{e_{ij}},\bm{e_{ik}})}}{\tau}
    ) } 
\end{equation}
\end{small}

\begin{equation}
    L_{ICL}=  \frac{1}{ \lvert Y_i \lvert } \sum_{Y_i} L_{ICL_j}
\end{equation}
 
\subsection{Training Objective}
To train the model, we combine the contrastive loss with cross-entropy and train them jointly. This aims to use a contrastive loss to close the distance between positive samples, while maximizing the probability of correct labels through a cross-entropy loss. The overall training objective is calculated as follows:
\begin{equation}
    L = \alpha \cdot L_{CL} + (1-\alpha) \cdot L_{BCE}
\end{equation}

where $ L_{CL} \in$ \{ \emph{SCL, ICL, JSCL, JSPCL, SLCL}\}.

\section{Experiments and Analysis}

\begin{table*}
\centering
\begin{tabular}{ccccc}
\hline
Info./Lang. & English & Arabic & Spanish & Indonesian\\
\hline
Train (\#) & 6,838 & 2,278 & 3,561 & 3373\\
Valid (\#) & 886 & 585 & 679 & 860 \\
Test (\#) & 3,259 & 1,518 & 2,854 & 1841\\
Total (\#) & 10,983 & 4,381 & 7,094 & 6074\\
Classes (\#) & 11 & 11 & 11 & 8\\
Type & MEC & MEC & MEC & MNC \\
\hline
\end{tabular}
\caption{Data Statistics.}
\label{tab:accents}
\end{table*}

\begin{table}[htbp]
\centering
\begin{tabular}{cccc}
\hline
Method & $F_{Macro}$ & $F_{Micro}$    & JS \\
\hline
BNN & 52.80 & 63.20 & - \\
ReRc & 53.90 & 65.10 & - \\
DATN & 55.10 & - & 58.30 \\
NTUA & 52.80 & 70.10 & 58.80 \\
LEM & 56.70 & 67.50 & - \\
\hline
SpanEmo & 57.00 & 70.32 & 58.30 \\
JSCL & \textbf{57.68} & \textbf{71.01} & \textbf{59.05} \\
JSPCL & 57.42 & 70.75 & 58.58 \\
SLCL & 56.62 & 70.9 & 58.9 \\
ICL & 57.59 & 70.49 & 58.6 \\
SCL & 57.63 & 70.8 & 58.89 \\
\hline
\end{tabular}
\caption{Experimental results on English dataset.}
\label{tab:accents}
\end{table}
 
\subsection{Dataset}
In order to investigate multi-label text classification tasks, we have selected the SemEval2018 \cite{mohammad2018semeval} multi-label emotion classification (MEC) task in English, Arabic, and Spanish as an illustrative example. The MEC datasets have been annotated to identify the presence of eleven discrete emotions, namely anger, anticipation, disgust, fear, joy, love, optimism, pessimism, sadness, surprise, and trust. In order to examine the efficacy and applicability of our approach, we have conducted experiments on a multi-label news classification (MNC) task in addition to the multi-label emotion classification task. For this purpose, we utilized an open source Indonesian multi-label news classification dataset \cite{10.1007/978-3-030-86331-9_44}, comprising 8 labels including society, politics, economy, technology, military, environment, culture, and others. Each sample in the dataset is associated with at most two category labels. The datasets were initially partitioned into three distinct subsets, namely the training set (Train), validation set (Valid), and test set (Test). For the purpose of training and testing, the default partitioning method of the dataset was directly employed. We evaluate our methods using the micro F1-score, macro F1-score, and Jaccard index score (JS) in accordance with the metrics in SemEval2018 \cite{mohammad2018semeval}. For each language, Table 1 summarizes the train, valid, and test sets and shows the number of instances in each set.
 
\subsection{Experimental Settings}
We use SpanEmo\footnote{Since our proposed method is based on SpanEmo for experiments, we also reproduce the experimental results of the method.} proposed by \citet{alhuzaliananiadou2021spanemo} as the base model. SpanEmo is a SOTA model for multi-label text classification tasks proposed by \citet{alhuzaliananiadou2021spanemo}, they trained the model with a loss combining the cross-entropy loss and the label-correlation aware (LCA) loss \cite{yeh2017learning}. We replaced the LCA loss of this model with several of our proposed contrastive losses for comparison. In addition to the SpanEmo model, we also compared the models with superior performance under each dataset separately. For the MEC task, the English models include JBNN \cite{10.1007/978-3-319-99495-6_21}, DATN \cite{yu2018improving}, NTUA \cite{baziotis2018ntua}, LEM \cite{Fei_Zhang_Ren_Ji_2020}, and ReRc \cite{zhou2018relevant}. On the Arabic dataset, we compare our method with EMA \cite{badaro2018ema}, Tw-StAR \cite{mulki2018tw}, HEF \cite{9007420} and BERT-base \cite{DBLP:journals/corr/abs-2008-09378}. On the Spanish dataset, we used Tw-StAR \cite{mulki-etal-2018-tw}, ELiRF \cite{gonzalez2018elirf}, MILAB \cite{mohammad2018semeval} and BERT-base \cite{DBLP:journals/corr/abs-2008-09378} as comparison models. To address the MNC task, we have identified and selected the state-of-the-art (SOTA) methods that have demonstrated superior performance on this dataset. The chosen methods comprise SGM \cite{yang2018sgm}, SU4MLC \cite{lin2018semanticunit}, mBERT \cite{DBLP:journals/corr/abs-2008-09378}, Indonesian-BERT \cite{10.1007/978-3-030-86331-9_44}, and Indonesian-BERT+Sim \cite{10.1007/978-3-030-86331-9_44}.

All experiments were carried out using PyTorch\footnote{https://pytorch.org/} and an RTX TITAN with 24 GB of memory. Using the open-source Hugging-Face implementation\footnote{https://huggingface.co/}, we fine-tuned “bert-base”\footnote{https://huggingface.co/bert-base-uncased} \cite{wolf2020transformers} for English. What's more, we selected “bert-base-arabic” \footnote{https://huggingface.co/asafaya/bert-base-arabic} constructed by \citet{safaya2020kuisail}. for Arabic and “bert-base-spanish-uncased”\footnote{https://huggingface.co/dccuchile/bert-base-spanish-wwm-uncased} constructed by \citet{canete2020spanish} for Spanish. We set the same hyper-parameters with a fixed initialization seed for three models training, where the batch size is 32 and the feature dimension is 768. The dropout rate is 0.1 and the early stop patience we set as 10 and 20 epochs. With a learning rate of 1e-3 for the FFN and 2e-5 for the BERT encoder, Adam was chosen for optimization. For the loss weight $\alpha$, we use the Hyperopt\footnote{http://hyperopt.github.io/hyperopt/} hyperparameter selection method \cite{10.5555/2986459.2986743} to search for the optimal parameters under each contrastive learning method. For each model, we used five different random seeds to carry out experiments, and the scores of five experiments were averaged as the final score.

 \begin{figure*}[htbp]
\centering
\addtocounter{figure}{-1}
\subfigure{
		\begin{minipage}[t]{0.32\linewidth}%%%%%%%%%note2
			\includegraphics[width=1\linewidth]{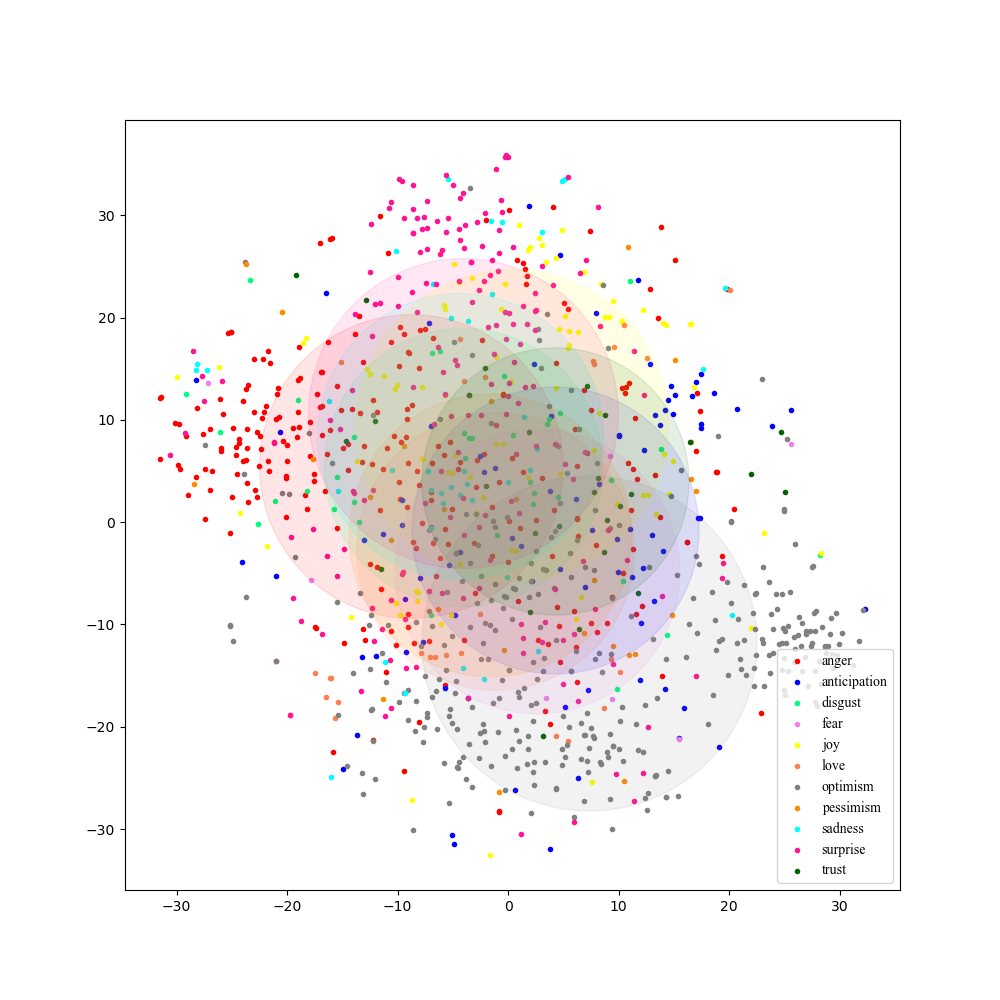}%%%%%%%%%note3
		\caption{2D visualization of \\ SpanEmo's semantic space.}
		\end{minipage}%
	}%
\subfigure{
		\begin{minipage}[t]{0.32\linewidth}%%%%%%%%%note2
			\includegraphics[width=1\linewidth]{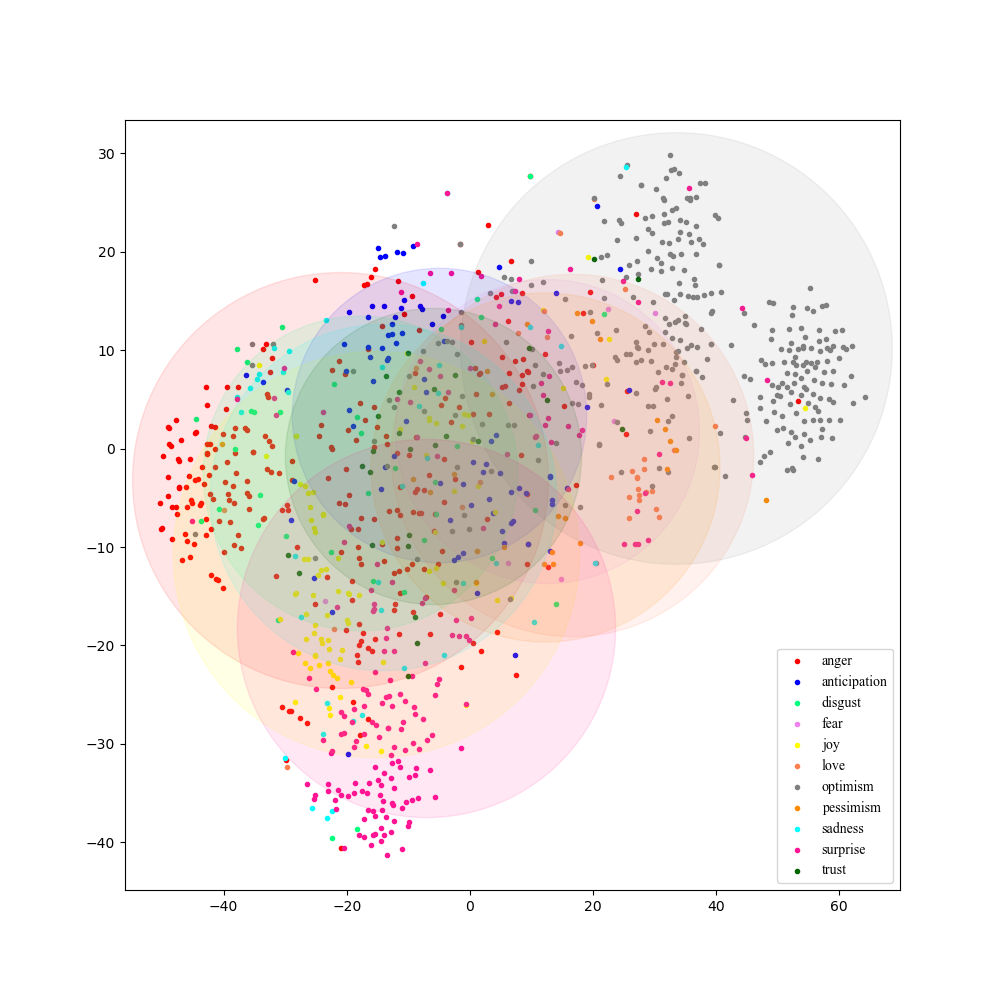}%%%%%%%%%note3
		\caption{2D visualization of \\ JSPCL's semantic space}
		\end{minipage}%
	}%
\subfigure{
		\begin{minipage}[t]{0.32\linewidth}%%%%%%%%%note2
			\includegraphics[width=1\linewidth]{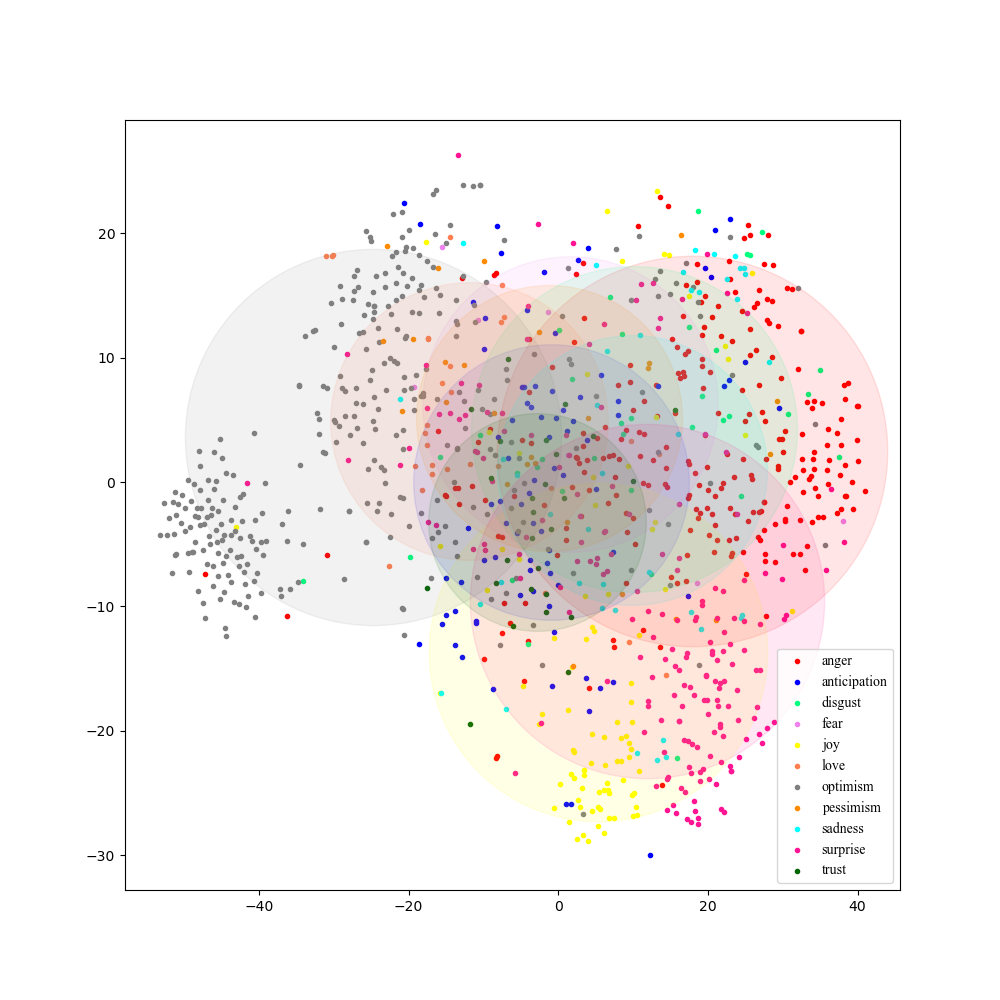}%%%%%%%%%note3
		\caption{2D visualization of \\ ICL's semantic space}
		\end{minipage}%
	}%
 \end{figure*}
 
\subsection{Results and Analysis}

\textbf{Main Performance for MEC.}  As shown in Table 2 to Table 4, all five of our contrastive learning strategies essentially delivered improvement to the model for the MEC task, with the JSCL approach performing best on the English dataset, reaching 57.68, 71.01 and 59.05 for $F_{Macro}$, $F_{Micro}$ and JS respectively, an improvement of 0.68, 0.69 and 0.75 over the SpanEmo model. The performance improvement of our method is more obvious on the Arabic dataset, where the $F_{Micro}$ value of the SLCL method is 1.25 higher than that of SpanEmo, and the $F_{Micro}$ and JS of the SCL method are improved by 0.90 and 1.49 respectively. The SCL method also performed well on the Spanish language dataset, achieving the highest JS value of 53.52. 

\textbf{Main Performance for MNC.} As shown in Table 5, in the task of MNC, the SCL method exhibits superior performance, achieving noteworthy scores of 74.29, 85.27, and 84.06 for $F_{Macro}$, $F_{Micro}$ and JS metrics respectively. These remarkable results substantiate the efficacy of the SCL approach in addressing the MNC challenge. Our five contrastive learning methods have a significant improvement effect on the model on $F_{Macro}$,, indicating that our methods can improve the categories with poor performance, thereby alleviating the class imbalance problem of MNC tasks to a certain extent.

\textbf{Comparison between Out-sample and In-sample.}  In general, one particular method, referred to as ICL, exhibits comparatively less improvement. This approach primarily emphasizes contrasting labels within a single sample, considering the labels present in the text as positive examples and those absent as negative examples. However, due to its limited ability to pay attention to label relationships across different texts, ICL fails to effectively capture the inherent distinctions among labels.

\textbf{Comparison between Strict Standard and Loose Standard.} Through the comparison between the loose standard loss JSCL and the strict standard loss SCL, we can find that the overall performance of SCL on the four datasets is better, that is, to a certain extent, strict standard contrastive learning methods are more suitable for multi-label text classification tasks than loose standard contrastive learning methods.

\begin{table}[htbp]
\centering
\begin{tabular}{cccc}
\hline
Method & $F_{Macro}$ & $F_{Micro}$    & JS \\
\hline
Tw-StAR & 44.60 & 59.70 & 46.50 \\
EMA & 46.10 & 61.80 & 48.90 \\
BERT$_{base}$ & 47.70 & 65.00 & 52.30 \\
HEF & 50.20 & 63.10 & 51.20 \\
\hline
SpanEmo & 53.63 & 65.81 & 53.94 \\
JSCL & 54.08 & 66.00 & 54.14 \\
JSPCL & 53.70 & 65.86 & 53.98 \\
SLCL & \textbf{54.88} & 66.37 & 54.65 \\
ICL & 54.26 & 66.13 & 54.17 \\
SCL & 54.27 & \textbf{66.71} & \textbf{55.43} \\
\hline
\end{tabular}
\caption{Experimental results on Arabic dataset.}
\label{tab:accents}
\end{table}

\textbf{Comparison between ProCL and FeaCL.} Through the results of JSCL and JSPCL, we could find that the method of the ProCL type does not perform as well as the method of the FeaCL type in terms of performance. We believe that because the semantic space of the multi-label text model is too complex, it is more effective to directly focus on the semantic space of the model than the probability distribution.

\begin{table}[htbp]
\centering
\begin{tabular}{cccc}
\hline
Method & $F_{Macro}$ & $F_{Micro}$    & JS \\
\hline
Tw-StAR & 39.20 & 52.00 & 43.80 \\
ELiRF & 44.00 & 53.50 & 45.80 \\
MILAB  & 40.70 & 55.80 & 46.90 \\
BERT$_{base}$ & 47.40 & 59.60 & 48.70 \\
\hline
SpanEmo & 55.49 & 63.34 & 52.68 \\
JSCL & 55.62 & 63.45 & 52.94 \\
JSPCL & \textbf{56.44} & \textbf{64.16} & 53.31 \\
SLCL & 56.00 & 63.56 & 52.69 \\
ICL & 55.82 & 63.46 & 52.66 \\
SCL & 55.88 & 63.70 & \textbf{53.52} \\
\hline
\end{tabular}
\caption{Experimental results on Spanish dataset.}
\label{tab:accents}
\end{table}

\begin{table}[htbp]
\centering
\begin{tabular}{cccc}
\hline
Method & $F_{Macro}$ & $F_{Micro}$    & JS \\
\hline
SGM & 44.08 & 74.24 & - \\
SU4MLC & 40.38 & 75.66 & - \\
mBERT & 66.56 & 81.85 & - \\
Indonesian-BERT & 67.57 & 84.53 & - \\
\makecell[c]{Indonesian-BERT \\ +Sim} & 70.82 & 84.66 & - \\
\hline
SpanEmo & 71.62 & 85.09 & 83.66 \\
JSCL & 73.13 & 85.23 & 83.98 \\
JSPCL & 72.17 & 84.91 & 82.81 \\
SLCL & 73.47 & 85.19 & 83.86 \\
ICL & 74.22 & 85.15 & 83.82 \\
SCL & \textbf{74.29} & \textbf{85.27} & \textbf{84.06} \\
\hline
\end{tabular}
\caption{Experimental results on Indonesian dataset.}
\label{tab:accents}
\end{table}

 \begin{figure*}[htbp]
\centering
\subfigure{
		\begin{minipage}[t]{0.32\linewidth}%%%%%%%%%note2
			\includegraphics[width=1\linewidth]{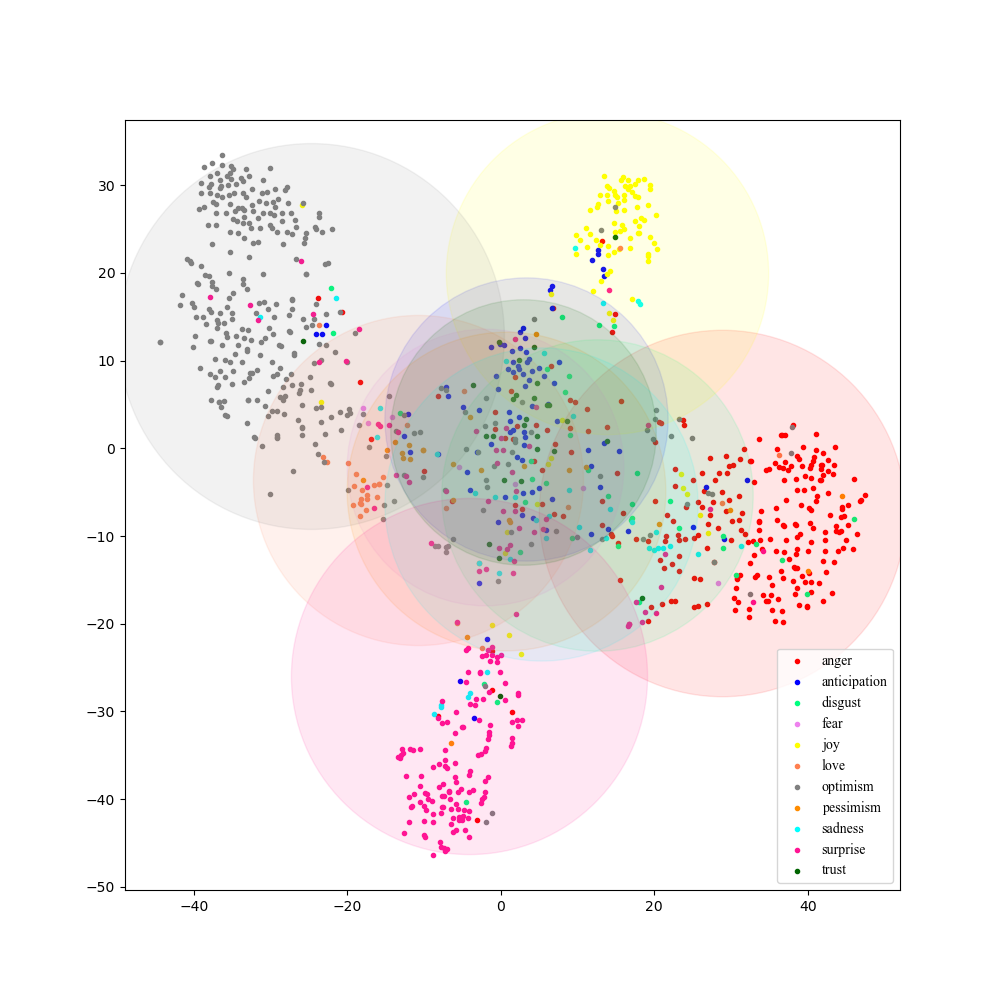}%%%%%%%%%note3
		\caption{2D visualization of \\ JSCL's semantic space}
		\end{minipage}%
	}%
 \subfigure{
		\begin{minipage}[t]{0.32\linewidth}%%%%%%%%%note2
			\includegraphics[width=1\linewidth]{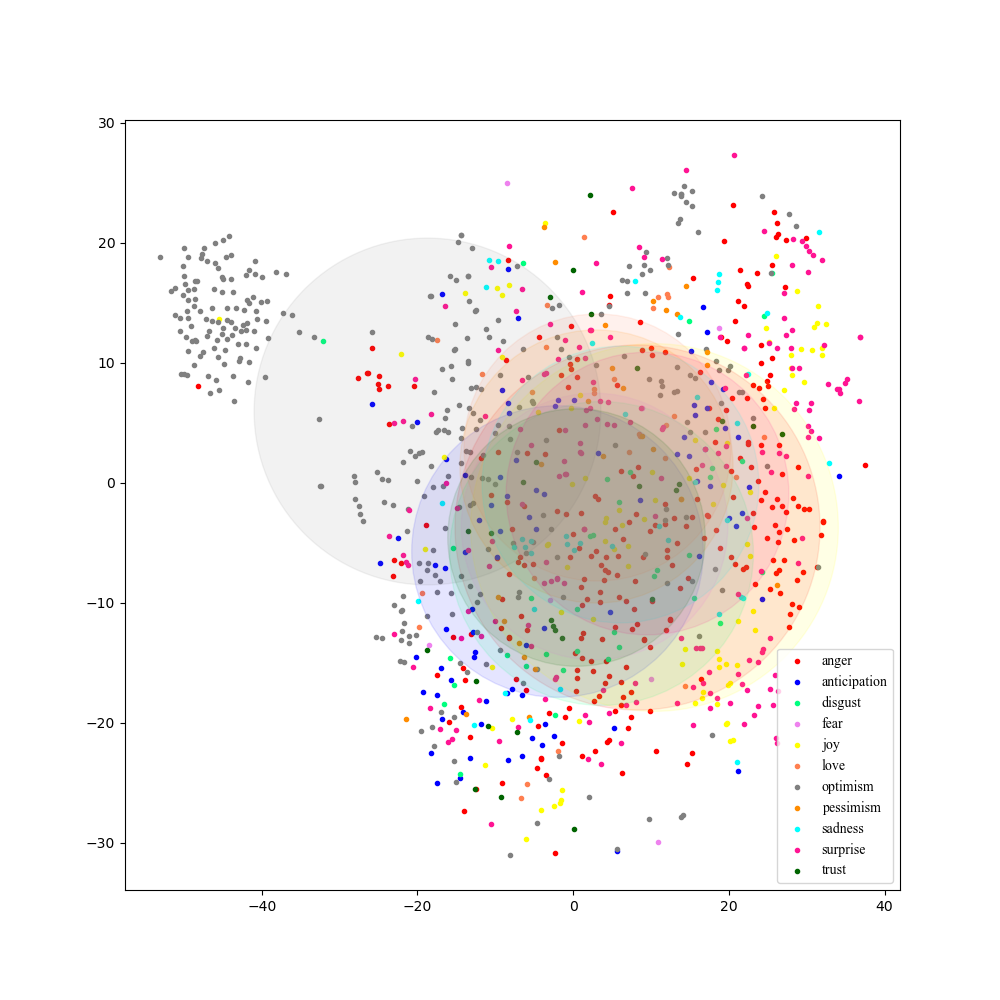}%%%%%%%%%note3
		\caption{2D visualization of \\ SLCL's semantic space}
		\end{minipage}%
	}%
  \subfigure{
		\begin{minipage}[t]{0.32\linewidth}%%%%%%%%%note2
			\includegraphics[width=1\linewidth]{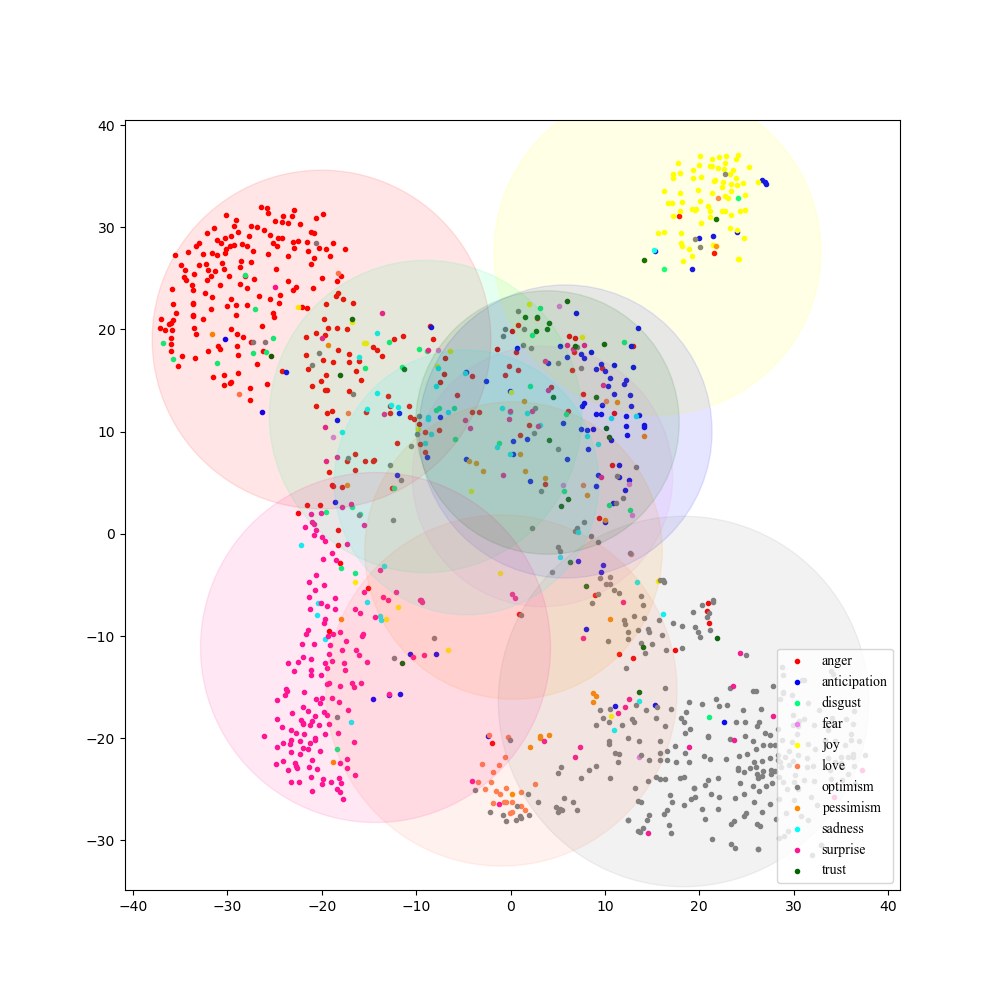}%%%%%%%%%note3
		\caption{2D visualization of \\ SCL's semantic space}
		\end{minipage}%
	}%
 \end{figure*}
 
\textbf{Interpretable Analysis.} Taking the experimental results in Spanish as an example, we analyze the interpretability of our method from the multi-label dimension and the single-label dimension respectively. In the multi-label dimension, we use the entire test set for analysis, consider the samples with identical labels to be under the same cluster, and then use the T-SNE method for dimensionality reduction and visualization. At the same time, we also calculated the Calinski-Harbasz score of cluster clustering to evaluate whether the semantic representation space of each category can be well discriminated. It is worth noting that under the single-label dimension, we only use the test set with only one label for interpretable analysis.

The interpretable analysis results for each method in the multi-label dimension and the single-label dimension are shown in Table 6. The larger the interpretable analysis results, the higher the discrimination of samples of different categories in the semantic space, and the better the semantic representation ability of the model. It can be seen that in addition to SLCL, other contrastive learning methods can make the samples of the same category in the semantic space more compact, and the boundaries of sample clusters of different categories are more obvious. SLCL aims to narrow the representation of categories, so it cannot make the boundaries between different categories more obvious. Among them, JSCL and SCL have better effects in optimizing the semantic representation space. As a rigorous contrastive learning method, SCL achieves the best results on multi-label dimension evaluation, with a Calinski-Harbasz value of 25.14. When evaluates from a multi-label perspective, JSCL performs slightly worse than SCL, but when evaluated from a single-label perspective, JSCL achieves the highest Calinski-Harbasz score of 200.48. We also further visualize the semantic space under the single-label dimension, as shown in Figures 3 to 8. It can be clearly seen that in JSCL and SCL, each category is more closely aggregated, and the boundaries among different categories are also more obvious.

 \begin{table}
\centering
\begin{tabular}{ccc}
\hline
Method & Multi-label & Singel-label \\
\hline
SpanEmo & 5.64 & 48.07 \\
JSCL & 24.07 & \textbf{200.48} \\
JSPCL & 17.80 & 131.54 \\
SLCL & 4.35 & 42.33 \\
ICL & 13.43 & 109.55 \\
SCL & \textbf{25.14} & 198.84 \\
\hline
\end{tabular}
\caption{Interpretable analysis results}
\label{tab:accents}
\end{table}

\section{Conclusion}
To investigate the efficacy of contrastive learning using various methodologies, we offer five effective contrastive losses for multi-label text classification tasks. The experimental results of this paper show that contrastive loss can improve the performance of multi-label text classification tasks. Furthermore, we find that strict criteria contrastive learning and feature-based contrastive learning outperform other contrastive learning methods on multi-label text classification tasks. In the future, based on these two methods, we will further explore the contrastive loss that is more suitable for multi-label text classification tasks.

% \section*{Ethics Statement}
% Scientific work published at ACL 2023 must comply with the ACL Ethics Policy.\footnote{\url{https://www.aclweb.org/portal/content/acl-code-ethics}} We encourage all authors to include an explicit ethics statement on the broader impact of the work, or other ethical considerations after the conclusion but before the references. The ethics statement will not count toward the page limit (8 pages for long, 4 pages for short papers).

\section*{Acknowledgements}
This work was supported by the Guangdong Basic and Applied Basic Research Foundation of China (No. 2023A1515012718).

\section*{Limitations}
This paper proposes five novel contrastive losses for multi-label text classification tasks. However, our method has the following limitations: 

1. We only selected the multi-label emotion classification task and  multi-label news classification as the representative of the multi-label text classification tasks. 

2. We only conduct experiments on the single modal of text, and have not extended to multi-modal tasks. 

3. Our method chooses the SpanEmo model as the backbone, lacking attempts to more models.
% Entries for the entire Anthology, followed by custom entries
\bibliography{anthology,custom}
\bibliographystyle{acl_natbib}

% \appendix

% \section{Example Appendix}
% \label{sec:appendix}

% This is a section in the appendix.

\end{document}